%% file: template.tex
\documentclass{article}

\usepackage{arxiv}

\usepackage[utf8]{inputenc} % allow utf-8 input
\usepackage[T1]{fontenc}    % use 8-bit T1 fonts
\usepackage{hyperref}       % hyperlinks
\usepackage{url}            % simple URL typesetting
\usepackage{booktabs}       % professional-quality tables
\usepackage{amsfonts}       % blackboard math symbols
\usepackage{nicefrac}       % compact symbols for 1/2, etc.
\usepackage{microtype}      % microtypography
\usepackage{lipsum}
\usepackage{graphicx}
\graphicspath{ {./images/} }

\usepackage{tabularx}
\usepackage{booktabs}
\usepackage{arydshln}

\usepackage{listings}

\usepackage[status=draft,nomargin,inline,lang=spanish]{fixme}
\usepackage{placeins}
\usepackage{xcolor}
\fxusetheme{colorsig}
\usepackage[table]{xcolor} % for colors
\usepackage{tcolorbox} % for colored boxes
\usepackage{array} % for defining column types
\usepackage{colortbl} % In your preamble
\usepackage{tikz}

\usepackage{soul}
\definecolor{lightorange}{RGB}{252, 229, 205}
\definecolor{lightblue}{RGB}{208, 224, 227}
\definecolor{lightyellow}{RGB}{255, 242, 204}
\definecolor{peachyhighlight}{RGB}{254, 236, 205}
\definecolor{pastelgreen}{RGB}{196, 223, 198}
\definecolor{lightpurple}{RGB}{224, 204, 255}
\definecolor{pastelpurple}{RGB}{205, 192, 230}
\definecolor{pastelpink}{RGB}{255, 221, 230}
\definecolor{lightgreen}{RGB}{204, 230, 204}
\definecolor{pastelpeach}{RGB}{255, 223, 204}
\definecolor{pastellavender}{RGB}{230, 215, 230}
\definecolor{pastelblue}{RGB}{174, 198, 207}

\newcolumntype{C}[1]{>{\centering\arraybackslash}p{#1}}

\tcbset{
  histstyle/.style={colback=lightorange, colframe=white, width=\linewidth, boxrule=0.4pt, sharp corners},
  repstyle/.style={colback=lightblue,   colframe=white, width=\linewidth, boxrule=0.4pt, sharp corners},
  measstyle/.style={colback=lightyellow, colframe=white, width=\linewidth, boxrule=0.4pt, sharp corners}
}

\title{Bias by Design? How Data Practices Shape Fairness in AI Healthcare Systems}

\author{
Anna Arias-Duart \\
    Orcid ID: 0000-0002-8819-6735 \\
    Barcelona Supercomputing Center (BSC) \\
  \texttt{anna.ariasduart@bsc.es} \\
   \And
 Maria Eugenia Cardello \\
Orcid ID: 0009-0004-6467-4511 \\
Barcelona Supercomputing Center (BSC) \\
  \texttt{maria.cardello@bsc.es} \\
  \And
 Atia Cortés \\
 Orcid ID: 0000-0002-4394-439X \\
Barcelona Supercomputing Center (BSC) \\
  \texttt{atia.cortes@bsc.es} \\
}

\begin{document}
\maketitle
\begin{abstract}
Artificial intelligence (AI) holds great promise for transforming healthcare. However, despite significant advances, the integration of AI solutions into real-world clinical practice remains limited. A major barrier is the quality and fairness of training data, which is often compromised by biased data collection practices. This paper draws on insights from the AI4HealthyAging project, part of Spain's national R\&D initiative, where our task was to detect biases during clinical data collection. We identify several types of bias across multiple use cases, including historical, representation, and measurement biases. These biases manifest in variables such as sex, gender, age, habitat, socioeconomic status, equipment, and labeling. We conclude with practical recommendations for improving the fairness and robustness of clinical problem design and data collection. We hope that our findings and experience contribute to guiding future projects in the development of fairer AI systems in healthcare. 
\end{abstract}

% keywords can be removed

\section{Introduction}\label{sec1}

The application of artificial intelligence (AI) in healthcare has grown rapidly in recent years, offering new possibilities for medical tasks such as diagnosis and clinical decision-making \cite{elhaddad2024ai, busnatu2022clinical, krishnan2023artificial}. Among the most transformative developments are the adoption of machine learning (ML), deep learning (DL) with the recent emergence of generative AI marking a significant new frontier. However, only a small fraction of these AI solutions are ultimately integrated into real-world healthcare systems \cite{gong2021nice}. This limited adoption can be attributed to different factors, including a mismatch with practical clinical needs or non-compliance with  the European Union’s AI Act \cite{aiact}, which emphasizes transparency, safety, and accountability in AI systems.

As a result, many hospitals are increasingly investing in the development of smaller, in-house models tailored to their specific clinical needs. However, training such models remains a significant challenge. Medical data is highly sensitive and typically subject to strict GDPR regulations \cite{EuropeanParliament2016a}, which limit access and use. Moreover, it is particularly hard to collect data, in terms of the time required to pass through the evaluation of protocols by ethical committees, the constraints of inclusion / exclusion criteria and the difficulty to find enough population aiming to participate into the study. Even so, the amount of collected data will often not be enough to train AI models. It does not get better with publicly available datasets, which often lack essential metadata, undermining both the generalizability and clinical relevance of trained models \cite{daneshjou2021lack}. 

Consequently, the successful implementation of in-house AI solutions often depends not on model architecture or performance metrics, but on a more fundamental, and frequently overlooked, factor: \textbf{how data collection for model training is defined and planned}. The success, reliability, and safety of AI systems in healthcare are deeply rooted in the quality, structure, and contextual appropriateness of the data on which they are built.

This paper draws on lessons learned from AI4HealthyAging, a national AI research project under Spain’s 2021 “R\&D Missions in Artificial Intelligence” Program, focused on developing AI solutions for age-related diseases. The project addressed a range of clinical use cases, including cardiovascular conditions, sarcopenia, sleep disorders, Parkinson’s disease, mental health, colorectal and prostate cancer, and hearing loss. Our work centered on identifying bias in data through stakeholder interviews and limited metadata analysis. Based on these insights, we present a set of recommendations to guide data planning and collection in clinical AI projects, aiming to improve fairness, quality, and reliability.

Before presenting the specific biases identified in our study, we begin in Section \ref{sec:bias_definition} by examining how bias is defined in the existing literature. Section \ref{sec:bias_categorization} explores how bias has been categorized in previous work. In Section \ref{sec:bias_in_practice}, we present the biases we identified, already categorized. Finally, Section \ref{sec:recommendations} offers a set of recommendations, measures that could have mitigated the biases observed and which we hope will support future clinical data collection efforts in AI development.

%\cite{chen2024internvl} %internvl

%PROBLEMA
%¿Como "definir / planificar" la recoleccion de datos para entrenar un modelo de IA?

%¿Qué variables debes plantearte para definir un estudio de IA en salud?

%¿Qué recomendaciones / variables se deben tener en cuenta?
%¿Qué diferencia hay entre un estudio "no medico" y un medico?
%¿Cual es la diferencia entre un estudio medico realizado por humanos, y un estudio con IA (correlaciones espureas)?
%¿Cada caso de uso definir una metodologia especifica, que debe contemplar el conocimiento experto en distintas etapas de la investigacion?

%Biases in medicine

%- Classification, Direction, and Prevention of Bias in Epidemiologic Research
%- 

%Paper de recomendaciones a tener en cuenta para recoger datos no sesgados en medicina.. 

%Pocas bases de datos abiertas. Las que hay no tienen metadatos. No se pueden reutilizar... 

\section{Bias definition} \label{sec:bias_definition}

Despite its widespread use, the term \texttt{bias} lacks a standardized definition in the AI field. Different subfields and applications interpret and operationalize bias in different ways depending on their specific context \cite{kordzadeh2022algorithmic, hanna2025ethical, koccak2025bias, panch2019artificial, olteanu2019social}. However, definitions across the literature tend to converge around three core elements: the source of the bias (\textit{e.g.,} algorithms, systems, or errors), how the bias manifests (\textit{e.g.,} through discrimination, unequal impacts, or prediction errors), and the entities affected by the bias (\textit{e.g.,} individuals, patients, or underrepresented groups). Table~\ref{tab:bias-definitions} presents a selection of definitions from the literature, spanning general AI to healthcare-specific contexts, organized and color-coded according to these three analytical dimensions.

These definitions of bias imply a normative judgment: that the outcomes it produces are undesirable, unjust, or detrimental to certain individuals or groups. This aligns closely with the concept of health equity, which is defined as the \textit{absence of systematic disparities in health (or in the major social determinants of health) between groups with different levels of underlying social advantage/disadvantage—that is, wealth, power, or prestige} \cite{braveman2003defining}. Understanding bias in AI as a normative issue highlights that biased outcomes are more than technical errors; they can reinforce social inequities, particularly in healthcare. Addressing bias is needed to achieving health equity by preventing unfair disparities.

\input{table_definitions}

\section{Bias categorization} \label{sec:bias_categorization}
Many studies in the literature have proposed different categorizations of the sources of bias in AI systems \cite{koccak2025bias, suresh2021framework, mehrabi2021survey, cross2024bias, chinta2024ai, schwartz2022towards, abramoff2023considerations, nazer2023bias, hasanzadeh2025bias, AGARWAL2023100702}. 
While terminology may vary across articles, most reflect a general consensus around three stages in the AI development pipeline where biases can originate: data, model development, and system implementation. Some works also propose additional stages that are also crucial to identify possible biases: \textit{(i)} an initial stage to formulate of the research problem, where the purpose, requirements and impact need to be evaluated \cite{koccak2025bias, abramoff2023considerations, nazer2023bias, hasanzadeh2025bias, AGARWAL2023100702}, and \textit{(ii)} a final phase of monitoring after deployment, which should be maintained as long as the AI system is in use \cite{abramoff2023considerations, hasanzadeh2025bias}. Table \ref{tab:bias-categorization} illustrates this alignment by mapping the terminology used in nine different papers to these five stages.

\input{categorization_table}

Although there is a broad agreement on the stages at which biases are introduced, the types of biases identified within these stages vary in both nomenclature and granularity across sources. Some categories are broad, such as \textit{selection bias} \cite{chinta2024ai}, while others are more fine-grained, like \textit{validity of the research question} \cite{koccak2025bias}. Certain labels serve as umbrella terms, for example \textit{representation bias} \cite{suresh2021framework}, under which more specific biases fall. For instance, \textit{demographic bias} \cite{koccak2025bias} can be considered a subcategory of representation bias. Furthermore, there is often overlap between categories, as certain biases span multiple dimensions. For example, \textit{institutional bias} \cite{koccak2025bias} can be understood as a combination of \textit{historical bias} \cite{suresh2021framework}, which reflects systemic inequalities, and \textit{aggregation bias} \cite{koccak2025bias}, where institutional practices rely on generalized data that may overlook the specific needs of marginalized groups.

\section{Bias in Practice} \label{sec:bias_in_practice}

%\begin{figure}[!t]
%\centering
%\includegraphics[width=0.95\linewidth]{img1.png}
%\caption{Categorization of identified biases.}\label{fig:categories}
%\end{figure}

In this section, we highlight several biases identified in our work that may affect the performance and generalizability of AI models. These biases arise from the first two steps detailed in the previous section: data design and data collection. To make these issues more tangible, we present concrete examples illustrating how such biases can be inadvertently introduced into training data, potentially compromising model fairness and validity.

For clarity, we categorize these biases according to the three sources of harm in data generation proposed by Suresh and Guttag \cite{suresh2021framework}: historical bias, representation bias, and measurement bias (see Table \ref{tab:categories}). As noted earlier, these categories are not mutually exclusive. For example, we classify gender bias as historical bias due to its roots in societal norms and systemic inequalities. However, it could also be considered a form of measurement bias if gender is inferred using subjective scoring methods, as the methodology itself can introduce additional bias.

\input{suresh_table}

\paragraph{\sethlcolor{lightorange}\hl{\textbf{Sex bias.}}} \label{par:sex}
 In the Parkinson’s study, the distribution of participants by age group and sex was generally balanced, except in the 40–49 and 80–89 age groups. The notably lower representation of females in the 80–89 group may be due to higher female mortality rates, which make recruiting females in this age range more difficult. These sex-based differences in participant distribution reflect important biological and disease-related factors. For example, research by Cerri et al. \cite{cerri2019parkinson} shows that males have about twice the risk of developing Parkinson’s disease compared to females, yet females tend to experience faster disease progression and higher mortality. Such sex differences in disease risk, progression, and survival underscore the importance of carefully considering sex as a key variable to avoid bias in data collection and analysis, ensuring predictive models accurately capture these nuances.

\paragraph{\sethlcolor{lightorange}\hl{\textbf{Gender bias.}}} \label{par:gender}
Although gender information was not directly available in the data we analyzed, a Gender Score could be derived as in \cite{yuan2021gender}. %That study showed hearing impairment is linked to cognitive decline, with individuals exhibiting masculine traits at greater risk, while those with feminine or androgynous traits may have a reduced risk. \acnote{are we relating hearing impairment to gender expression?} 
Gender bias is particularly important to consider in healthcare contexts. For instance, Samulowitz et al. \cite{samulowitz2018brave} demonstrated that gender norms influence pain treatment: women with pain received less effective relief, fewer opioid prescriptions, more antidepressants, and more mental health referrals compared to men. Neglecting gender data and its proper analysis can exacerbate existing inequalities and lead to biased health outcomes.

\paragraph{\sethlcolor{lightblue}\hl{\textbf{Age bias.}}} \label{par:age}

Because the project focuses on age-related conditions, the control group, composed of participants without the disease, tent to be younger on average, while the disease groups include older individuals. This difference arises because these diseases primarily affect older adults, making it easier to recruit younger healthy controls but harder to find older participants without the condition. %\acnote{is this in general or in a specific use case? not sure if we have this recorded but I think often control group were relatives of the disease group} \aanote{I meant in general, but if we want to add an specific case we saw that in the Alzheimer study}

Another example of age bias is found in the Parkinson’s study, where the majority of subjects were between 60 and 79 years old. This aligns well with the known prevalence of Parkinson’s, which affects approximately 3\% of people at age 65 and up to 5\% of those over 85 \cite{dexter2013parkinson}. Additionally, the median age increased with disease severity, 64 years for severity 1, 71.5 years for severity 2, and 75 years for severity 3, with no participants younger than 60 in the most severe category. This further reflects the strong association between age and disease progression. However, such uneven age distributions can introduce age bias in AI models trained on this data. Hence, models may learn to associate age-related features with disease presence or severity rather than true disease-specific markers.

\paragraph{\sethlcolor{lightblue}\hl{\textbf{Habitat bias.}}}\label{par:habitat}

This bias arises when geographic or environmental context affects participant representation. In this project, most participants came from urban areas, largely because the hospitals conducting the studies were located in urban settings. Travel distance and accessibility can be significant barriers for individuals living in rural areas, making it less likely for them to participate or remain involved in long-term studies. Even in urban areas, usually hospitals concentrate patients from some specific regions of the city that are determined by socio-economic factors or environmental exposure that have an impact on their quality of life \cite{ijerph19116745}, which leads to the following type of bias.

\paragraph{\sethlcolor{lightblue}\hl{\textbf{Socioeconomic bias.}}} \label{par:socioeconomic}

It occurs when participants’ social and economic factors, such as income, education, occupation, or access to healthcare, influence who is included in a study. For example, in one of the studies, data was collected from a private hospital. Because private hospitals typically serve patients with higher income levels or better insurance coverage, this creates a socioeconomic bias by primarily including individuals from wealthier backgrounds. 

In the hearing loss study, control group participants tended to have higher education levels than other groups. Education often correlates with quieter work environments, while lower education may correspond to noisier jobs (\textit{e.g.}, factory work). Ignoring these factors could lead to misleading conclusions about hearing loss causes.

\paragraph{\sethlcolor{lightyellow}\hl{\textbf{Equipment bias.}}} \label{par:equipment}

It occurs when variations in the devices used for data collection, such as different models, calibration settings, or software, affect measurement consistency. This can lead to results that are not comparable across participants or sites.
For example, in the hearing loss study, most participants had cochlear implants from the same manufacturer. As a result, findings on quality of life and cognitive improvement may not generalize to users of other implant types, potentially biasing the model toward the characteristics of one specific device.

\paragraph{\sethlcolor{lightyellow}\hl{\textbf{Labeling bias.}}} \label{par:labeling}

This bias occurs when data labels, such as diagnoses or classifications, are influenced by human judgment or local context. This can occur when different people use inconsistent criteria, or even when the same team labels all data but follows specific institutional practices. For example, labels from one hospital with its unique diagnostic style may not generalize well elsewhere, reducing model accuracy and fairness in real-world settings.

An example of labeling bias was found in the hearing loss study, specifically in the classification of participants’ occupations. Initially, the dataset used standardized occupational categories\footnote{\url{https://www.ine.es/dyngs/INEbase/operacion.htm?c=Estadistica_C&cid=1254736177033&menu=enlaces&idp=1254735976614}}
 that did not include a significant group of society: homemakers. After adding this category, it was revealed that 35\% of women fell into this group, with no male representation. This initial omission and subsequent reclassification demonstrate how labeling categories influenced by human decisions can misrepresent certain groups.

\paragraph{\sethlcolor{peachyhighlight}\hl{\textbf{Intersectional bias.}}} \label{par:intersectional}

It occurs when two or more demographic variables interact in a way that affects the fairness or validity of a model. In the Alzheimer’s study, a potential intersectional bias involving age and sex was observed across three diagnostic groups: control, mild cognitive impairment (MCI), and Alzheimer’s. On average, female were younger than male across all groups: the age difference is two years in both the control and Alzheimer’s groups, and four years in the MCI group.
%This could be particularly concerning in light of prior research \cite{han2024aging}, which found that biological sex significantly influences brain function as measured by EEG, but that these sex-based differences only emerge in older adults (ages 30–80), not in younger ones (ages 20–30). As a result, the younger average age of female in the study may lead to an underrepresentation of sex-related neurophysiological effects in the female subgroup, potentially introducing bias into the predictive models. 
If the interaction between age and sex is not properly controlled, models may misattribute these normative age-related sex differences to disease-specific changes, compromising the validity and fairness of diagnostic predictions.

\section{Recommendations}\label{sec:recommendations}

In this section, we present recommendations for mitigating bias in medical data collection. We organize them using the same three categories as before: historical, representation, and measurement biases. These categories are not strictly separated, some recommendations may address multiple types of bias.

\paragraph{\sethlcolor{lightorange}\hl{Historical bias.}}

\begin{itemize}

\item Involve a diverse, \textbf{interdisciplinary group} in planning the experiment. Collection design may be influenced by the implicit biases of those responsible for data collection. As research has consistently shown, healthcare providers often exhibit biases toward %marginalized minorities
historically %and intentionally 
excluded groups \cite{fitzgerald2017implicit, maina2018decade, sabin2015health}, and these biases are likely to persist in the absence of curricula specifically focused on minority health~\cite{phelan2019effects}. Furthermore, stakeholders hold divergent views on the nature, significance, and mitigation of bias in healthcare AI \cite{aquino2025practical}. As such, assembling a diverse and interdisciplinary team is important to incorporate multiple perspectives, minimize bias, and ensure that data collection strategies are equitable and inclusive.

\item Ensure that data is collected in an \textbf{aggregated or disaggregated} manner when appropriate. As Cirillo et al. \cite{cirillo2020sex} explain, bias can be desirable or undesirable. Including sex and gender, for example, may improve prediction accuracy in cardiovascular diseases \cite{dhruva2011gender}, but may also reinforce harmful assumptions, such as higher reported depression rates among women \cite{martin2013experience}. It is therefore essential to review existing literature and carefully plan what metadata to collect and use, to avoid unintended harm.

%\cite{greenwald1995implicit}
\end{itemize}

\paragraph{\sethlcolor{lightblue}\hl{\textbf{Representation bias.}}}

\begin{itemize}
    \item Define clear and balanced \textbf{inclusion and exclusion criteria}. Criteria should be specific but not overly restrictive, to maintain sample diversity and enable the formation of appropriate control groups. In this project, the study population was older, which made it challenging to find age-matched control groups for those with the comorbidity. As a result, the control groups had lower mean values, potentially introducing spurious correlations and biasing the results.
    \item Analyse the need to include an \textbf{intersectional benchmark} to better represent the targeted population \cite{buolamwini18a}. This will help refining the evaluation metrics and understand the health condition. 
    \item Ensure the \textbf{sample size} is feasible and sustainable. This involves assessing recruitment and retention potential within time and resources constraints. Engage experts with experience in similar studies to identify potential challenges, such as high dropout rates or participant burden. For instance, in this study, some protocols had to be shortened, as their extended duration was too demanding for participants. 
\end{itemize}

\paragraph{\sethlcolor{lightyellow}\hl{\textbf{Equipment bias.}}} 

\begin{itemize}
    \item Evaluate the \textbf{data labeling} process. Review how data has been labeled to ensure that categories are clear and consistent. Well-defined labeling reduces ambiguity, improves data quality becoming an essential step to align with FAIR principles \cite{FAIR}. Depending on the type of data, labeling should be approached differently, for example, socioeconomic variables may benefit from input by an interdisciplinary team, while clinical related data should not be labeled by a single professional alone, in order to minimize personal bias. 
    
    \item Consider \textbf{equipment} and deployment context. It is important to account for the equipment used during data collection and where the model will ultimately be deployed. If both data collection and deployment occur within the same hospital using the same equipment, consistency is maintained. However, if a model is trained on data from one type of equipment and then applied to data form another, equipment-related bias may arise. This can compromise the model's performance and limit its generalizability.
\end{itemize}

%\begin{itemize}

%\item Representatividad

%\item Posibles sesgos en la enfermedad

%\item Sesgos históricos (dar ejemplos)

%\begin{itemize}
%    \item Género
%    \item Raza
%    \item Socioeconómico
%\end{itemize}

%\item Sesgo de medición 
%\begin{itemize}
%    \item Sesgo de etiquetado
%\end{itemize}
%\end{itemize}

\section{Conclusions}

To successfully incoporate AI systems into healthcare, clinical AI projects must address bias not only as a technical issue but also as a matter of governance. Our work highlights how different forms of bias can emerge during data collection, illustrated with real cases from our project. We provide a list of recommendations to avoid these biases and emphasize the importance of interdisciplinary collaboration, balanced cohort design, and thoughtful inclusion of metadata. We hope that these lessons learned from our experience will inform and support future healthcare AI projets in building more equitable and effective systems that are both legally compliant and socially responsible.

\section*{Acknowledgments}

We would like to thank all the collaborators involved in the project who participated in the interviews, and Amparo Callejón-Leblic for her insightful feedback. This research has been funded by the Artificial Intelligence for Healthy Aging (AI4HA, MIA.2021.M02.0007.E03) project from the Programa Misiones de I+D en Inteligencia Artificial 2021 and by the European Union-NextGenerationEU, Ministry of Universities and Recovery, Transformation and Resilience Plan, through a call from Universitat Polit\`ecnica de Catalunya and Barcelona Supercomputing Center (Grant Ref. 2021UPC-MS-67461/2021BSC-MS-67461). Anna Arias Duart acknowledges her AI4S fellowship within the “Generación D” initiative by \href{https://www.red.es/es}{Red.es}, Ministerio para la Transformación Digital y de la Función Pública, for talent attraction (C005/24-ED CV1), funded by NextGenerationEU through PRTR. Additional funding from the European Union through the Marie Skłodowska-Curie project AHEAD (grant agreement No 101183031). We would also like to thank Nardine Osman and Mark d’Inverno for Figure 2 in their work \cite{osman2024modelling}, which inspired our Table~\ref{tab:bias-definitions}.

%\end{acknowledgments}

%% The declaration on generative AI comes in effect
%% https://ceur-ws.org/GenAI/Policy.html
\section*{Declaration on Generative AI}

During the preparation of this work, the authors used ChatGPT (GPT-4) and DeepSeek Chat for grammar and spelling checks. After using these tools, the authors reviewed and edited the content as needed and take full responsibility for the publication’s content.

\bibliographystyle{unsrt}  
\bibliography{sample-ceur}  %%% Remove comment to use the external .bib file (using bibtex).


\end{document}

%% file: table_definitions.tex
\setlength{\dashlinedash}{0.5pt}   % Length of each dash
\setlength{\dashlinegap}{1.5pt}    % Gap between dashes
\setlength{\arrayrulewidth}{0.2pt} % Thickness of dash line

\begin{table}[htb]
\caption{Different definitions of bias and algorithmic bias in AI, ranging from general to healthcare-specific contexts. Highlights indicate the source of bias (\sethlcolor{lightgreen}\hl{green}), how it manifests (\sethlcolor{pastellavender}\hl{purple}), and the affected groups or outcomes (\sethlcolor{lightyellow}\hl{yellow}).}

\centering
\renewcommand{\arraystretch}{1.15} % slightly less vertical padding
\begin{tabularx}{\textwidth}{@{}l p{1.8cm} X@{}}
\toprule
\textbf{Context} & \textbf{Term} & \textbf{Definition} \\
\midrule

General & Algorithmic \newline bias
& It occurs when the outputs of an \sethlcolor{lightgreen}\hl{algorithm} benefit or disadvantage certain \sethlcolor{lightyellow}\hl{individuals} or \sethlcolor{lightyellow}\hl{groups} more than others without a justified reason for such \sethlcolor{pastellavender}\hl{unequal impacts} \cite{kordzadeh2022algorithmic}. \\
\hdashline

General & Bias  
& It refers to systematic and \sethlcolor{pastellavender}\hl{unfair favoritism} or \sethlcolor{pastellavender}\hl{prejudice} in AI \sethlcolor{lightgreen}\hl{systems}, which can lead to \sethlcolor{lightyellow}\hl{discriminatory outcomes} \cite{hanna2025ethical}.\\
\hdashline

%General & Risk of bias
%& It refers to the potential for a \sethlcolor{lightgreen}\hl{systematic error} (bias) in the estimators of the model’s predictive performance for the \sethlcolor{lightyellow}\hl{target population} or \sethlcolor{lightyellow}\hl{populations of interest}. Bias can act in either direction, with potential for \sethlcolor{pastellavender}\hl{overestimation} or \sethlcolor{pastellavender}\hl{underestimation} of the true model performance \cite{moons2025probast}. \\
%\hdashline

Health & Algorithmic \newline bias
& The instances when the application of an \sethlcolor{lightgreen}\hl{algorithm} compounds existing inequities in \sethlcolor{lightyellow}\hl{socioeconomic status, race, ethnic background, religion, gender, disability or sexual orientation} to \sethlcolor{pastellavender}\hl{amplify} them and adversely impact \sethlcolor{pastellavender}\hl{inequities} in health systems \cite{panch2019artificial}. \\
\hdashline

Health & Bias 
& It refers to \sethlcolor{lightgreen}\hl{systematic errors} leading to a \sethlcolor{pastellavender}\hl{distance} between \sethlcolor{pastellavender}\hl{prediction} and \sethlcolor{pastellavender}\hl{truth}, to the potential detriment of \sethlcolor{lightyellow}\hl{all} or \sethlcolor{lightyellow}\hl{some patients} \cite{koccak2025bias}. \\

\bottomrule
\end{tabularx}
\label{tab:bias-definitions}
\end{table}

%% file: categorization_table.tex
\begin{table}[h!]
\centering
\caption{Comparison of sources of bias with the AI/ML system life cycle.}
\renewcommand{\arraystretch}{1.5}
\begin{tabular}{>{\centering\arraybackslash}m{2.5cm} >{\centering\arraybackslash}m{2.5cm} >{\centering\arraybackslash}m{2.5cm} >{\centering\arraybackslash}m{2.5cm} >{\centering\arraybackslash}m{2.5cm} :>{\centering\arraybackslash}m{1cm}}
\hline
\parbox[c][0.7cm][t]{0.5cm}{\vspace{0.1cm}\includegraphics[height=0.5cm]{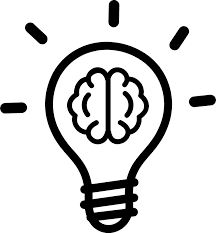}} &
\parbox[c][0.7cm][t]{0.5cm}{\vspace{0.1cm}\includegraphics[height=0.5cm]{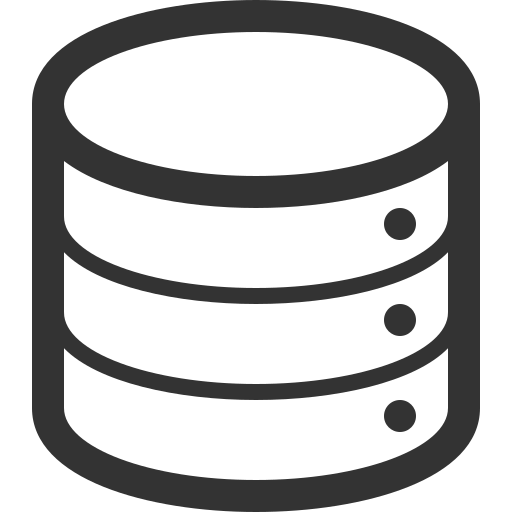}} &
\parbox[c][0.7cm][t]{0.5cm}{\vspace{0.1cm}\includegraphics[height=0.5cm]{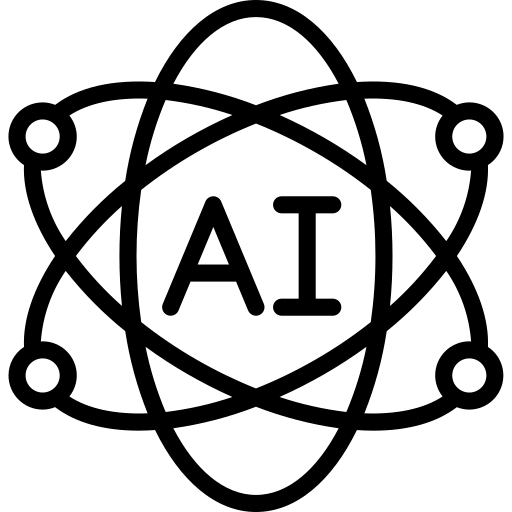}} &
\parbox[c][0.7cm][t]{0.5cm}{\vspace{0.1cm}\includegraphics[height=0.5cm]{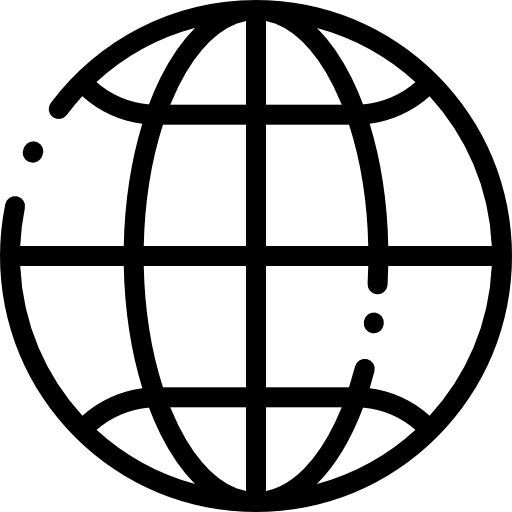}} &
\parbox[c][0.7cm][t]{0.5cm}{\vspace{0.1cm}\includegraphics[height=0.5cm]{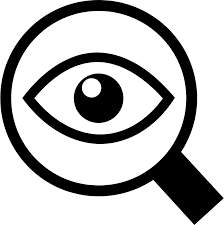}} &
\parbox[c][0.7cm][t]{0.5cm}{\vspace{0.1cm}\includegraphics[height=0.5cm]{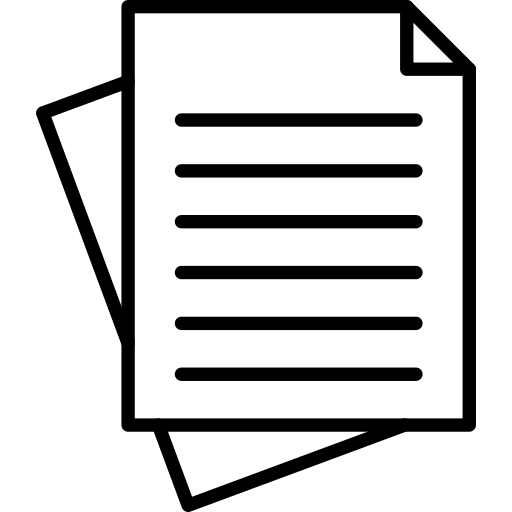}}\\
\hline
Design & Data & Modeling & Deployment & - & \cite{koccak2025bias} \\
- & Data Generation & Model Building & Implementation & - & \cite{suresh2021framework} \\
- & Data & Algorithm & User Interaction & - & \cite{mehrabi2021survey} \\
- & Training Data / Publication & Model Development \& Evaluation & Model Implementation & - & \cite{cross2024bias} \\
- & Pre-processing & In-processing & Post-processing & - & \cite{chinta2024ai} \\
Conception \& Design & Development & Validation & Access & Monitoring & \cite{abramoff2023considerations} \\
Formulating Research Problem & Data Collection \& Pre-processing & Model Development \& Validation & Model Implementation & - & \cite{nazer2023bias} \\
Conception & Data Collection \& Pre-processing & In-processing & Post-processing & Post-deployment Surveillance & \cite{hasanzadeh2025bias} \\
Problem scope & Data used & Model building & Decisions supported by analytical tool & - & \cite{AGARWAL2023100702} \\
\hline
\end{tabular}
\label{tab:bias-categorization}
\end{table}

%% file: suresh_table.tex
%\begin{table}[h!]
%\centering
%\caption{Identified biases categorized according to the classification proposed by Suresh and Guttag~\cite{suresh2021framework}. Each bias listed is clickable and links to a detailed explanation in the text.}
%\vspace{1ex}
%\begin{tabular}{C{0.3\linewidth} C{0.3\linewidth} C{0.3\linewidth}}

%\multicolumn{3}{c}{\textbf{Problem Design and Data Collection}} \\
%\addlinespace[1ex]
%\toprule
%\sethlcolor{lightorange}\hl{Historical Biases} & \sethlcolor{lightblue}\hl{Representation Biases} & \sethlcolor{lightyellow}\hl{Measurement Biases}\\
%\midrule

%\hyperref[par:sex]{Sex} & 
%\hyperref[par:age]{Age} & 
%\hyperref[par:equipment]{Equipment} \\

%\hyperref[par:gender]{Gender} & 
%\hyperref[par:habitat]{Habitat} & 
%\hyperref[par:labeling]{Labeling} \\

%~ & 
%\hyperref[par:socioeconomic]{Socioeconomic} & 
%~ \\
%\bottomrule

%\end{tabular}
%\label{tab:categories}
%\end{table}

\begin{table}[h!]
\centering
\caption{Identified biases categorized according to the classification proposed by Suresh and Guttag~\cite{suresh2021framework}. Each bias listed is clickable and links to a detailed explanation in the text.}
\vspace{1ex}
\begin{tabular}{C{0.3\linewidth} C{0.3\linewidth} C{0.3\linewidth}}

\multicolumn{3}{c}{\textbf{Problem Design and Data Collection}} \\
\addlinespace[1ex]
\toprule
\tcbox[histstyle]{Historical Biases} &
\tcbox[repstyle]{Representation Biases} &
\tcbox[measstyle]{Measurement Biases} \\
%\midrule

\hyperref[par:sex]{Sex} &
\hyperref[par:age]{Age} &
\hyperref[par:equipment]{Equipment} \\

\hyperref[par:gender]{Gender} &
\hyperref[par:habitat]{Habitat} &
\hyperref[par:labeling]{Labeling} \\

~ &
\hyperref[par:socioeconomic]{Socioeconomic} &
~ \\
\bottomrule

\end{tabular}
\label{tab:categories}
\end{table}